\documentclass{article}

\usepackage{microtype}
\usepackage{graphicx}
\usepackage{subfigure}
\usepackage{booktabs} 
\usepackage[utf8]{inputenc} 
\usepackage[T1]{fontenc}    
\usepackage{url}            
\usepackage{amsfonts}       
\usepackage{nicefrac}       
\usepackage{multirow}
\usepackage{mathrsfs}
\usepackage{amsmath}
\usepackage{amssymb}
\usepackage{comment}
\usepackage{bbm}
\usepackage{hyperref}
\usepackage{courier}

\usepackage[accepted]{icml2018}

\icmltitlerunning{PredRNN++: Towards A Resolution of the Deep-in-Time Dilemma in Spatiotemporal Predictive Learning}

\begin{document}

\twocolumn[

\icmltitle{PredRNN++: Towards A Resolution of the Deep-in-Time Dilemma in Spatiotemporal Predictive Learning}

\begin{icmlauthorlist}
\icmlauthor{Yunbo Wang}{tsinghua}
\icmlauthor{Zhifeng Gao}{tsinghua}
\icmlauthor{Mingsheng Long}{tsinghua}
\icmlauthor{Jianmin Wang}{tsinghua}
\icmlauthor{Philip S. Yu}{tsinghua}
\end{icmlauthorlist}

\icmlaffiliation{tsinghua}{School of Software, Tsinghua University, Beijing, China. E-mail: wangyb15@mails.tsinghua.edu.cn}

\icmlcorrespondingauthor{Mingsheng Long}{mingsheng@tsinghua.edu.cn}

\icmlkeywords{Predictive learning, Spatiotemporal learning, Deep learning}

\vskip 0.3in
]

\printAffiliationsAndNotice{}

\begin{abstract}

We present \textit{PredRNN++}, a recurrent network for spatiotemporal predictive learning. In pursuit of a great modeling capability for short-term video dynamics, we make our network deeper in time by leveraging a new recurrent structure named \textit{Causal LSTM} with cascaded dual memories. To alleviate the gradient propagation difficulties in deep predictive models, we propose a \textit{Gradient Highway Unit}, which provides alternative quick routes for the gradient flows from outputs back to long-range previous inputs. The gradient highway units work seamlessly with the causal LSTMs, enabling our model to capture the short-term and the long-term video dependencies adaptively. Our model achieves state-of-the-art prediction results on both synthetic and real video datasets, showing its power in modeling entangled motions.

\end{abstract}

\section{Introduction}

Spatiotemporal predictive learning is to learn the features from label-free video data in a self-supervised manner (sometimes called unsupervised) and use them to perform a specific task. This learning paradigm has benefited or could potentially benefit practical applications, e.g. precipitation forecasting \cite{shi2015convolutional,wang2017predrnn}, traffic flows prediction \cite{zhang2017deep,xu2018predcnn} and physical interactions simulation \cite{lerer2016learning,Finn2016Unsupervised}. 

An accurate predictive learning method requires effectively modeling video dynamics in different time scales. Consider two typical situations: (i) When sudden changes happen, future images should be generated upon nearby frames rather than distant frames, which requires that the predictive model learns short-term video dynamics; (ii) When the moving objects in the scene are frequently entangled, it would be hard to separate them in the generated frames. This requires that the predictive model recalls previous contexts before the occlusion happens. Thus, video relations in the short term and the long term should be adaptively taken into account.


\begin{figure*}[h]
\vskip 0.05in
\centering
\subfigure[Stacked ConvLSTMs]{
\includegraphics[width=0.65\columnwidth]{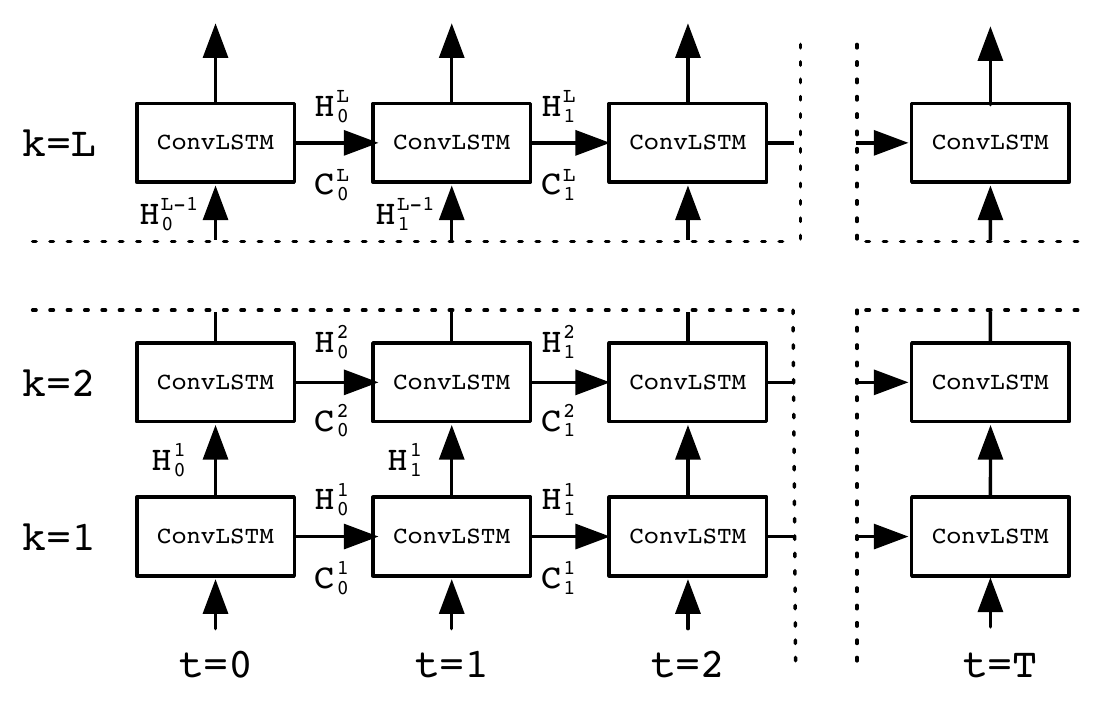}
\label{fig:stacked_convlstm}
}
\subfigure[Deep Transition ConvLSTMs]{
\includegraphics[width=0.65\columnwidth]{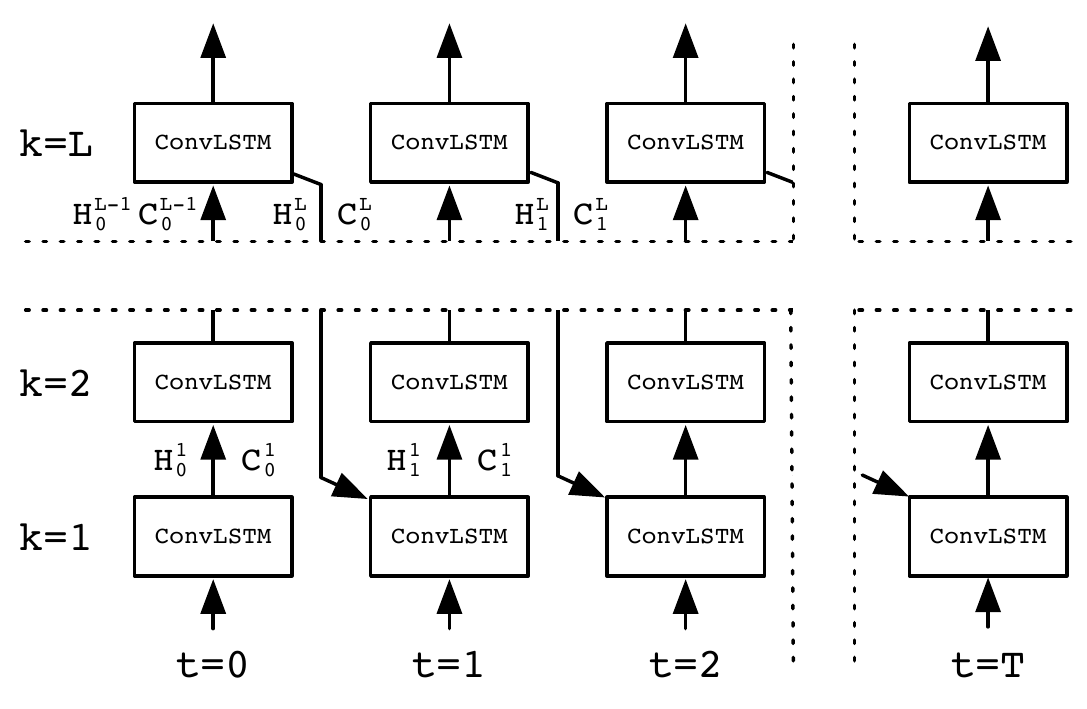}
\label{fig:deeplstm}
}
\subfigure[PredRNN with ST-LSTMs]{
\includegraphics[width=0.65\columnwidth]{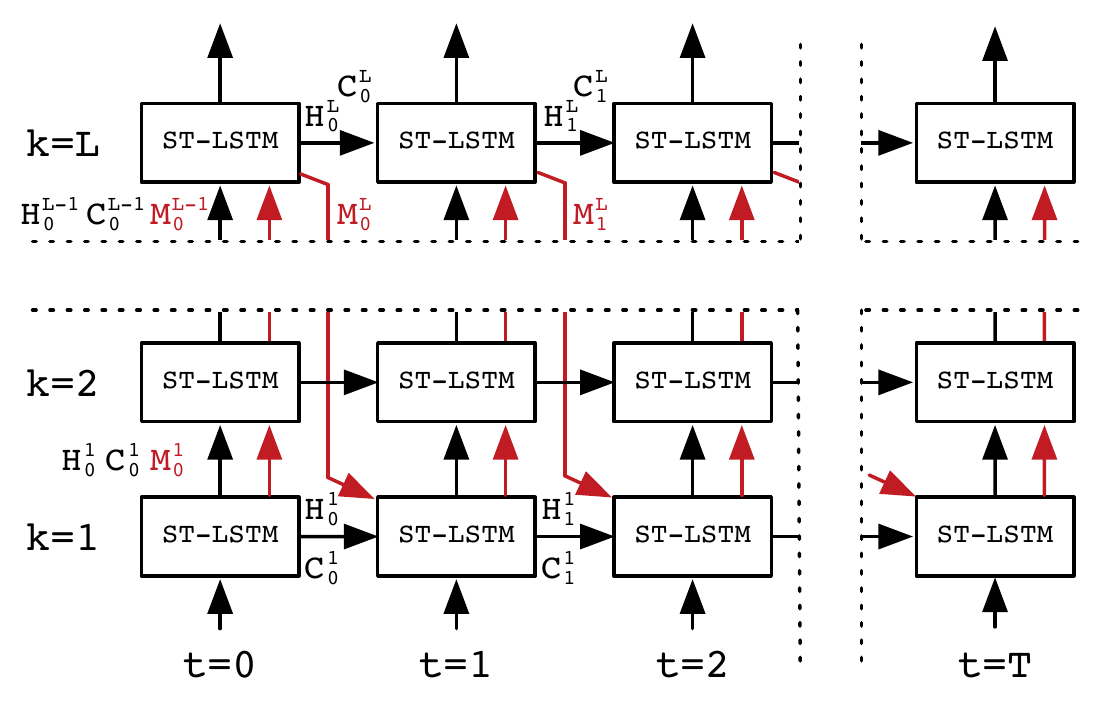}
\label{fig:predrnn}
}
\caption{Comparison of data flows in (a) the stacked ConvLSTM network, (b) the deep transition ConvLSTM network, and (c) PredRNN with the spatiotemporal LSTM (ST-LSTM). The two memories of PredRNN work in parallel: the red lines in subplot (c) denote the deep transition paths of the spatial memory, while horizontal black arrows indicate the update directions of the temporal memories.}
\label{fig:model_comparison}
\end{figure*}

\subsection{Deep-in-Time Structures and Vanishing Gradients Dilemma in Spatiotemporal Modeling}

In order to capture the long-term frame dependencies, recurrent neural networks (RNNs) \cite{rumelhart1988learning,werbos1990backpropagation,williams1995gradient} have been recently applied to video predictive learning \cite{Ranzato2014Video}. However, most methods \cite{srivastava2015unsupervised,shi2015convolutional,patraucean2015spatio} followed the traditional RNNs chain structure and did not fully utilize the network depth. The transitions between adjacent RNN states from one time step to the next are modeled by simple functions, though theoretical evidence shows that deeper networks can be exponentially more efficient in both spatial feature extraction \cite{bianchini2014complexity} and sequence modeling \cite{Gustavsson2012On}. We believe that making the network deeper-in-time, i.e. increasing the number of recurrent states from the input to the output, would significantly increase its strength in learning short-term video dynamics.

Motivated by this, a former state-of-the-art model named PredRNN \cite{wang2017predrnn} applied complex nonlinear transition functions from one frame to the next, constructing a dual memory structure upon Long Short-Term Memory (LSTM) \cite{hochreiter1997long}. Unfortunately, this complex structure easily suffers from the vanishing gradient problem \cite{bengio1994learning,Gustavsson2012On}, that the magnitude of the gradients decays exponentially during the back-propagation through time (BPTT). There is a \textbf{dilemma} in spatiotemporal predictive learning: the increasingly deep-in-time networks have been designed for complex video dynamics, while also introducing more difficulties in gradients propagation. Therefore, how to maintain a steady flow of gradients in a deep-in-time predictive model, is a path worth exploring. Our key insight is to build adaptive connections among RNN states or layers, providing our model with both longer routes and shorter routes at the same time, from input frames to the expected future predictions.

\section{Related Work} 

Recurrent neural networks (RNNs) are widely used in video prediction.
\citet{Ranzato2014Video} constructed a RNN model to predict the next frames.
\citet{srivastava2015unsupervised} adapted the sequence to sequence LSTM framework for multiple frames prediction.
\citet{shi2015convolutional} extended this model and presented the convolutional LSTM (ConvLSTM) by plugging the convolutional operations in recurrent connections.
\citet{Finn2016Unsupervised} developed an action-conditioned predictive model that explicitly predicts a distribution over pixel motion from previous frames.
\citet{Lotter2016Deep} built a predictive model upon ConvLSTMs, mainly focusing on increasing the prediction quality of the next one frame.
\citet{Villegas2017Decomposing} proposed a network that separates the information components (motion and content) into different encoder pathways.
\citet{patraucean2015spatio} predicted intermediate pixel flow and applied the flow to predict image pixels.
\citet{Kalchbrenner2016Video} proposed a sophisticated model combining gated CNN and ConvLSTM structures. It estimates pixel values in a video one-by-one using the well-established but complicated PixelCNNs \cite{van2016conditional}, thus severely suffers from low prediction efficiency.
\citet{wang2017predrnn} proposed a deep-transition RNN with two memory cells, where the spatiotemporal memory flows through all RNN states across different RNN layers.

Convolutional neural networks (CNNs) are also involved in video prediction, although they only create representations for fixed size inputs.
\citet{Oh2015Action} defined a CNN-based autoencoder model for Atari games prediction.
\citet{de2016dynamic} adapted filter operations of the convolutional network to the specific input samples.
\citet{villegas2017learning} proposed a three-stage framework with additional annotated human joints data to make longer predictions. 

To deal with the inherent diversity of future predictions, \citet{babaeizadeh2017stochastic} and \citet{denton2018stochastic} explored stochastic variational methods in video predictive models. But it is difficult to assess the performance of these stochastic models. Generative adversarial networks \cite{Goodfellow2014Generative,Denton2015Deep} were employed to video prediction \cite{Mathieu2015Deep,vondrick2016generating,bhattacharjee2017temporal,denton2017unsupervised,lu2017flexible,tulyakov2017mocogan}. These methods attempt to preserve the sharpness of the generated images by treating it as a major characteristic to distinguish real/fake video frames. But the performance of these models significantly depends on a careful training of the unstable adversarial networks. 

In summary, prior video prediction models yield different drawbacks. CNN-based approaches predict a limited number of frames in one pass. They focus on spatial appearances rather than the temporal coherence in long-term motions. The RNN-based approaches, in contrast, capture temporal dynamics with recurrent connections. However, their predictions suffer from the well-known vanishing gradient problem of RNNs, thus particularly rely on closest frames. In our preliminary experiments, it was hard to preserve the shapes of the moving objects in generated future frames, especially after they overlapped. In this paper, we solve this problem by proposing a new gradient highway recurrent unit, which absorbs knowledge from previous video frames and effectively leverages long-term information.

\section{Revisiting Deep-in-Time Architectures}


A general method to increase the depth of RNNs is stacking multiple hidden layers. A typical stacked recurrent network for video prediction \cite{shi2015convolutional} can be presented as Figure \ref{fig:stacked_convlstm}. The recurrent unit, ConvLSTM, is designed to properly keep and forget past information via gated structures, and then fuse it with current spatial representations. Nevertheless, stacked ConvLSTMs do not add extra modeling capability to the step-to-step recurrent state transitions.


In our preliminary observations, increasing the step-to-step transition depth in ConvLSTMs can significantly improve its modeling capability to the short-term dynamics. As shown in Figure \ref{fig:deeplstm}, the hidden state, $\mathcal{H}$, and memory state, $\mathcal{C}$, are updated in a zigzag direction. The extended recurrence depth between horizontally adjacent states enables the network to learn complex non-linear transition functions of nearby frames in a short interval. However, it introduces vanishing gradient issues, making it difficult to capture long-term correlations from the video. Though a simplified cell structure, the recurrent highway \cite{zilly2016recurrent}, might somewhat ease this problem, it sacrifices the spatiotemporal modeling power, exactly as the dilemma described earlier.



Based on the deep transition architecture, a well-performed predictive learning approach, PredRNN \cite{wang2017predrnn}, added extra connections between adjacent time steps in a stacked spatiotemporal LSTM (ST-LSTM), in pursuit of both long-term coherence and short-term recurrence depth. Figure \ref{fig:predrnn} illustrates its information flows. PredRNN leverages a dual memory mechanism and combines, by a simple concatenation with gates, the horizontally updated temporal memory $\mathcal{C}$ with the vertically transformed spatial memory $\mathcal{M}$. Despite the favorable information flows provided by the spatiotemporal memory, this parallel memory structure followed by a concatenation operator, and a $1\times1$ convolution layer for a constant number of channels, is not an efficient mechanism for increasing the recurrence depth. Besides, as a straight-forward combination of the stacked recurrent network and the deep transition network, PredRNN still faces the same vanishing gradient problem as previous models.




\section{PredRNN++}
In this section, we would give detailed descriptions of the improved predictive recurrent neural network (PredRNN++). Compared with the above deep-in-time recurrent architectures, our approach has two key insights: First, it presents a new spatiotemporal memory mechanism, causal LSTM, in order to increase the recurrence depth from one time step to the next, and by this means, derives a more powerful modeling capability to stronger spatial correlations and short-term dynamics. Second, it attempts to solve gradient back-propagation issues for the sake of long-term video modeling. It constructs an alternative gradient highway, a shorter route from future outputs back to distant inputs. 

\subsection{Causal LSTM}

\begin{figure}[htb]
\vskip 0.15in
\centering
\includegraphics[width=\columnwidth]{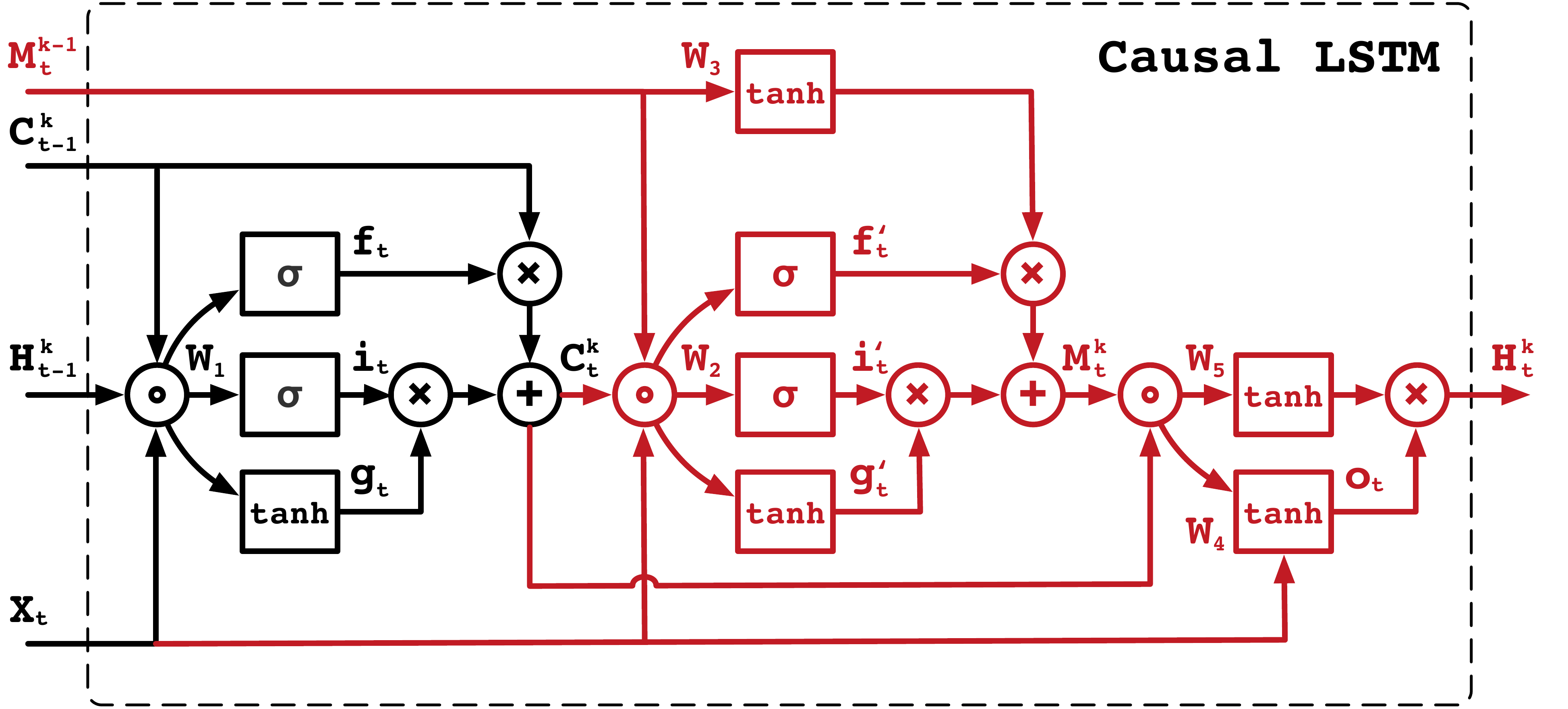}
\caption{Causal LSTM, in which the temporal and spatial memories are connected in a cascaded way through gated structures. Colored parts are newly designed operations, concentric circles denote concatenation, and $\sigma$ is the element-wise Sigmoid function.}
\label{fig:causal_lstm}
\end{figure}

The causal LSTM is enlightened by the idea of adding more non-linear layers to recurrent transitions, increasing the network depth from one state to the next. A schematic of this new recurrent unit is shown in Figure \ref{fig:causal_lstm}. A causal LSTM unit contains dual memories, the temporal memory $\mathcal{C}_t^k$, and the spatial memory $\mathcal{M}_t^k$, where the subscript $t$ denotes the time step, while the superscript denotes the $k^\text{th}$ hidden layer in a stacked causal LSTM network. The current temporal memory directly depends on its previous state $\mathcal{C}_{t-1}^k$, and is controlled through a forget gate $f_t$, an input gate $i_t$, and an input modulation gate $g_t$. The current spatial memory $\mathcal{M}_t^k$ depends on $\mathcal{M}_t^{k-1}$ in the deep transition path.  Specifically for the bottom layer ($k=1$), we assign the topmost spatial memory at $(t-1)$ to $\mathcal{M}_t^{k-1}$. Evidently different from the original spatiotemporal LSTM \cite{wang2017predrnn}, causal LSTM adopts a cascaded mechanism, where the spatial memory is particularly a function of the temporal memory via another set of gate structures. Update equations of the causal LSTM at the $k^\text{th}$ layer can be presented as follows:

\begin{equation}\label{equ:clstm}
\begin{split}
&\left(\begin{aligned} g_t \\ i_t \\ f_t \\ \end{aligned}\right) =\left(\begin{aligned} \tanh \\ \sigma \\ \sigma \\ \end{aligned}\right) W_1 \ast \left[\mathcal{X}_t, \mathcal{H}_{t-1}^k, \mathcal{C}_{t-1}^k \right] \\
&\mathcal{C}_t^k=f_t \odot \mathcal{C}_{t-1}^k + i_t \odot g_t \\ 
&\left(\begin{aligned} g_t^\prime \\ i_t^\prime \\ f_t^\prime \\ \end{aligned}\right) = \left(\begin{aligned} \tanh \\ \sigma \\ \sigma \\ \end{aligned}\right) W_2 \ast \left[\mathcal{X}_t, \mathcal{C}_t^k, \mathcal{M}_t^{k-1} \right] \\
&\mathcal{M}_t^k = f_t^\prime \odot \tanh\left(W_3 \ast \mathcal{M}_t^{k-1}\right) + i_t^\prime \odot g_t^\prime \\
&o_t = \tanh \left(W_4 \ast \left[\mathcal{X}_t, \mathcal{C}_t^k, \mathcal{M}_t^k\right]\right)\\
&\mathcal{H}_t^k = o_t \odot \tanh \left( W_5 \ast \left[\mathcal{C}_t^k, \mathcal{M}_t^k\right]\right)\\
\end{split} 
\end{equation}

where $\ast$ is convolution, $\odot$ is the element-wise multiplication, $\sigma$ is the element-wise Sigmoid function, the square brackets indicate a concatenation of the tensors and the round brackets indicate a system of equations. $W_{1\sim 5}$ are convolutional filters, where $W_3$ and $W_5$ are $1\times1$ convolutional filters for changing the number of filters. The final output $\mathcal{H}_t^k$ is co-determined by the dual memory states $\mathcal{M}_t^k$ and $\mathcal{C}_t^k$.

Due to a significant increase in the recurrence depth along the spatiotemporal transition pathway, this newly designed cascaded memory is superior to the simple concatenation structure of the spatiotemporal LSTM \cite{wang2017predrnn}. Each pixel in the final generated frame  would have a larger receptive field of the input volume at every time step, which endows the predictive model with greater modeling power for short-term video dynamics and sudden changes.  

We also consider another spatial-to-temporal causal LSTM variant. We swap the positions of the two memories, updating $\mathcal{M}_t^k$ in the first place, and then calculating $\mathcal{C}_t^k$ based on $\mathcal{M}_t^k$. An experimental comparison of these two alternative structures would be presented in Section \ref{sec:exp}, in which we would demonstrate that both of them lead to better video prediction results than the original spatiotemporal LSTM.

\subsection{Gradient Highway}

Beyond short-term video dynamics, causal LSTMs tend to suffer from gradient back-propagation difficulties for the long term. In particular, the temporal memory $\mathcal{C}_t^k$ may forget the outdated frame appearance due to longer transitions. Such a recurrent architecture remains unsettled, especially for videos with periodic motions or frequent occlusions. We need an information highway to learn skip-frame relations. 

\begin{figure}[t]
\vskip 0.15in
\centering
\includegraphics[width=\columnwidth]{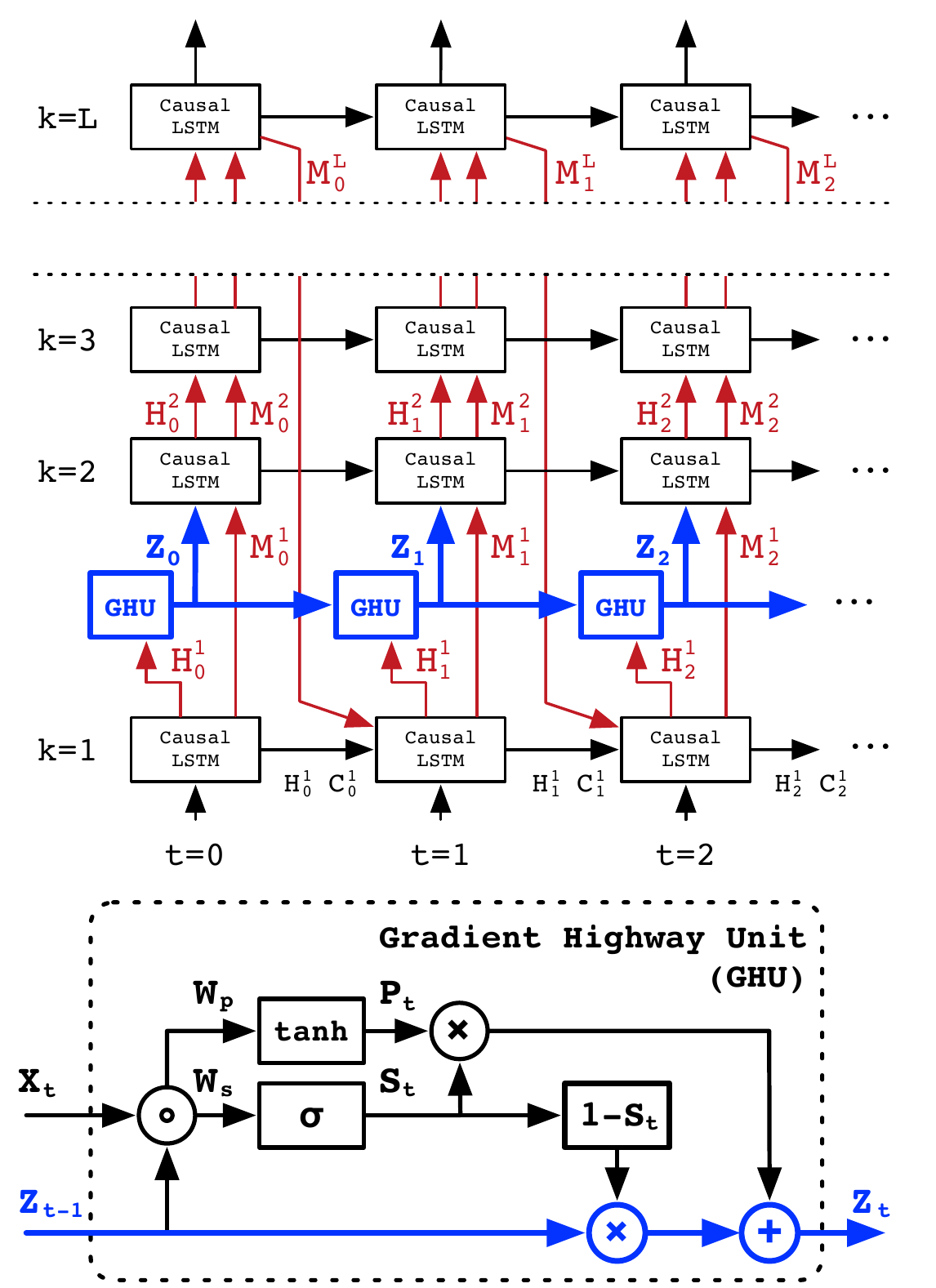}
\caption{Final architecture (top) with the gradient highway unit (bottom), where concentric circles denote concatenation, and $\sigma$ is the element-wise Sigmoid function. Blue parts indicate the gradient highway connecting the current time step directly with previous inputs, while the red parts show the deep transition pathway.}
\label{fig:gradient_highway}
\end{figure}

Theoretical evidence indicates that highway layers \cite{srivastava2015training} are able to deliver gradients efficiently in very deep feed-forward networks. We exploit this idea to recurrent networks for keeping long-term gradients from quickly vanishing, and propose a new spatiotemporal recurrent structure named \textit{Gradient Highway Unit} (GHU), with a schematic shown in Figure~\ref{fig:gradient_highway}. Equations of the GHU can be presented as follows: 

\begin{equation}\label{equ:GHU}
\small
\begin{split}
\mathcal{P}_t &= \tanh\left(W_{px} \ast \mathcal{X}_t + W_{pz} \ast \mathcal{Z}_{t-1}\right) \\
\mathcal{S}_t &= \sigma\left(W_{sx} \ast \mathcal{X}_t + W_{sz} \ast \mathcal{Z}_{t-1}\right) \\
\mathcal{Z}_t &= \mathcal{S}_t \odot \mathcal{P}_t + (1-\mathcal{S}_t) \odot \mathcal{Z}_{t-1} \\ 
\end{split} 
\end{equation}

where $W_{\bullet\bullet}$ stands for the convolutional filters. $\mathcal{S}_t$ is named as \textit{Switch Gate}, since it enables an adaptive learning between the transformed inputs $\mathcal{P}_t$ and the hidden states $\mathcal{Z}_{t-1}$. Equation \ref{equ:GHU} can be briefly expressed as $\mathcal{Z}_t = \texttt{GHU}(\mathcal{X}_t,\mathcal{Z}_{t-1})$.

In pursuit of great spatiotemporal modeling capability, we build a deeper-in-time network with causal LSTMs, and then attempt to deal with the vanishing gradient problem with the GHU. The final architecture is shown in Figure \ref{fig:gradient_highway}. Specifically, we stack $L$ causal LSTMs  and inject a GHU between the $1^\text{st}$ and the $2^\text{nd}$ causal LSTMs. Key equations of the entire model are presented as follows (for $3\le k \leq L$):

\begin{equation}\label{equ:architecture}
\small
\begin{split}
\mathcal{H}_t^1,\mathcal{C}_t^1,\mathcal{M}_t^1 &= \texttt{CausalLSTM}_\texttt{1}(\mathcal{X}_t,\mathcal{H}_{t-1}^1,\mathcal{C}_{t-1}^1,\mathcal{M}_{t-1}^L) \\
\mathcal{Z}_t &= \texttt{GHU}(\mathcal{H}_t^1,\mathcal{Z}_{t-1}) \\
\mathcal{H}_t^2,\mathcal{C}_t^2,\mathcal{M}_t^2 &= \texttt{CausalLSTM}_\texttt{2}(\mathcal{Z}_t,\mathcal{H}_{t-1}^2,\mathcal{C}_{t-1}^2,\mathcal{M}_t^1) \\
\mathcal{H}_t^k,\mathcal{C}_t^k,\mathcal{M}_t^k &= \texttt{CausalLSTM}_\texttt{k}(\mathcal{H}_t^{k-1},\mathcal{H}_{t-1}^k,\mathcal{C}_{t-1}^k,\mathcal{M}_t^{k-1}) \\
\end{split} 
\end{equation}

In this architecture, the gradient highway works seamlessly with the causal LSTMs to separately capture long-term and short-term video dependencies. With quickly updated hidden states $\mathcal{Z}_t$, the gradient highway shows an alternative quick route from the very first to the last time step (the blue line in Figure \ref{fig:gradient_highway}). But unlike temporal skip connections, it controls the proportions of $\mathcal{Z}_{t-1}$ and the deep transition features $\mathcal{H}_{t}^1$ through the switch gate $\mathcal{S}_t$, enabling an adaptive learning of the long-term and the short-term frame relations. 

We also explore other architecture variants by injecting GHU into a different hidden layer slot, for example, between the $(L-1)^\text{th}$ and $L^\text{th}$ causal LSTMs. Experimental comparisons would be given in Section \ref{sec:exp}. The network discussed above outperforms the others, indicating the importance of modeling characteristics of raw inputs rather than the abstracted representations at higher layers.


As for network details, we observe that the numbers of the hidden state channels, especially those in lower layers, have strong impacts on the final prediction performance. We thus propose a 5-layer architecture, in pursuit of high prediction quality with reasonable training time and memory usage, consisting of 4 causal LSTMs with 128, 64, 64, 64 channels respectively, as well as a 128-channel gradient highway unit on the top of the bottom causal LSTM layer. We also set the convolution filter size to $5$ inside all recurrent units.

\section{Experiments}
\label{sec:exp}

\begin{table*}[tb]
\caption{Results of PredRNN++ comparing with other models. We report per-frame SSIM and MSE of generated sequences. Higher SSIM or lower MSE denotes higher prediction quality. (*) indicates models that are not open source and are reproduced by us or others.}
\label{tab:mnist_mse}
\vskip 0.15in
\centering
\begin{small}
\begin{sc}
\renewcommand{\multirowsetup}{\centering}  
\begin{tabular}{lcccccc}
\toprule
\multirow{3}{0cm}{Model} & \multicolumn{4}{c}{MNIST-2} & \multicolumn{2}{c}{MNIST-3} \\
& \multicolumn{2}{c}{10 time steps} & \multicolumn{2}{c}{30 time steps} & \multicolumn{2}{c}{10 time steps} \\
& SSIM & MSE & SSIM & MSE & SSIM & MSE  \\
\midrule
FC-LSTM \cite{srivastava2015unsupervised} & 0.690 & 118.3 &  0.583 & 180.1 & 0.651 & 162.4\\
ConvLSTM \cite{shi2015convolutional} & 0.707 & 103.3 & 0.597 & 156.2 & 0.673 &142.1 \\
TrajGRU \cite{shi2017deep} & 0.713 & 106.9 & 0.588 & 163.0 & 0.682 & 134.0 \\
CDNA \cite{Finn2016Unsupervised} & 0.721 & 97.4 & 0.609 & 142.3 & 0.669 &138.2 \\
DFN \cite{de2016dynamic} & 0.726 & 89.0 & 0.601 & 149.5 & 0.679 & 140.5 \\
VPN* \cite{Kalchbrenner2016Video} & 0.870 & 64.1 & 0.620 & 129.6 & 0.734 & 112.3 \\
PredRNN \cite{wang2017predrnn} & 0.867 & 56.8 & 0.645 & 112.2 & 0.782 & 93.4 \\
\midrule
Causal LSTM & 0.882 & 52.5 & 0.685 & 100.7 & 0.795 & 89.2  \\
Causal LSTM (Variant: spatial-to-temporal) & 0.875 & 54.0 & 0.672 & 103.6 & 0.784 & 91.8 \\
PredRNN + GHU & 0.886 & 50.7 & 0.713 & 98.4 & 0.790 & 88.9 \\
Causal LSTM + GHU (Final) & \textbf{0.898} & \textbf{46.5} & \textbf{0.733} & \textbf{91.1} & \textbf{0.814} & \textbf{81.7} \\
\bottomrule
\end{tabular}
\end{sc}
\end{small}
\end{table*}

To measure the performance of our approach, we use two video prediction datasets in this paper: a synthetic dataset with moving digits and a real video dataset with human actions. For codes and results on more datasets, please refer to \url{https://github.com/Yunbo426/predrnn-pp}. 

We train all compared models using TensorFlow \cite{Abadi2016TensorFlow} and optimize them to convergence using ADAM \cite{Kingma2014Adam} with a starting learning rate of $10^{-3}$. Besides, we apply the scheduled sampling strategy \cite{bengio2015scheduled} to all of the models to stitch the discrepancy between training and inference. As for the objective function, we use the $L1$ + $L2$ loss to simultaneously enhance the sharpness and the smoothness of the generated frames.

\subsection{Moving MNIST Dataset}

\paragraph{Implementation}
We first follow the typical setups on the Moving MNIST dataset by predicting 10 future frames given 10 previous frames. Then we extend the predicting time horizon from 10 to 30 time steps to explore the capability of the compared models in making long-range predictions.
Each frame contains 2 handwritten digits bouncing inside a $64 \times 64$ grid of image. 
To assure the trained model has never seen the digits during inference period, we sample digits from different parts of the original MNIST dataset to construct our training set and test set. The dataset volume is fixed, with $10,000$ sequences for the training set, $3,000$ sequences for the validation set and $5,000$ sequences for the test set. 
In order to measure the generalization and transfer ability, we evaluate all models trained with $2$ moving digits on another $3$ digits test set.

\begin{figure}[htb]
\vskip 0.15in
\centering
\includegraphics[width=\columnwidth]{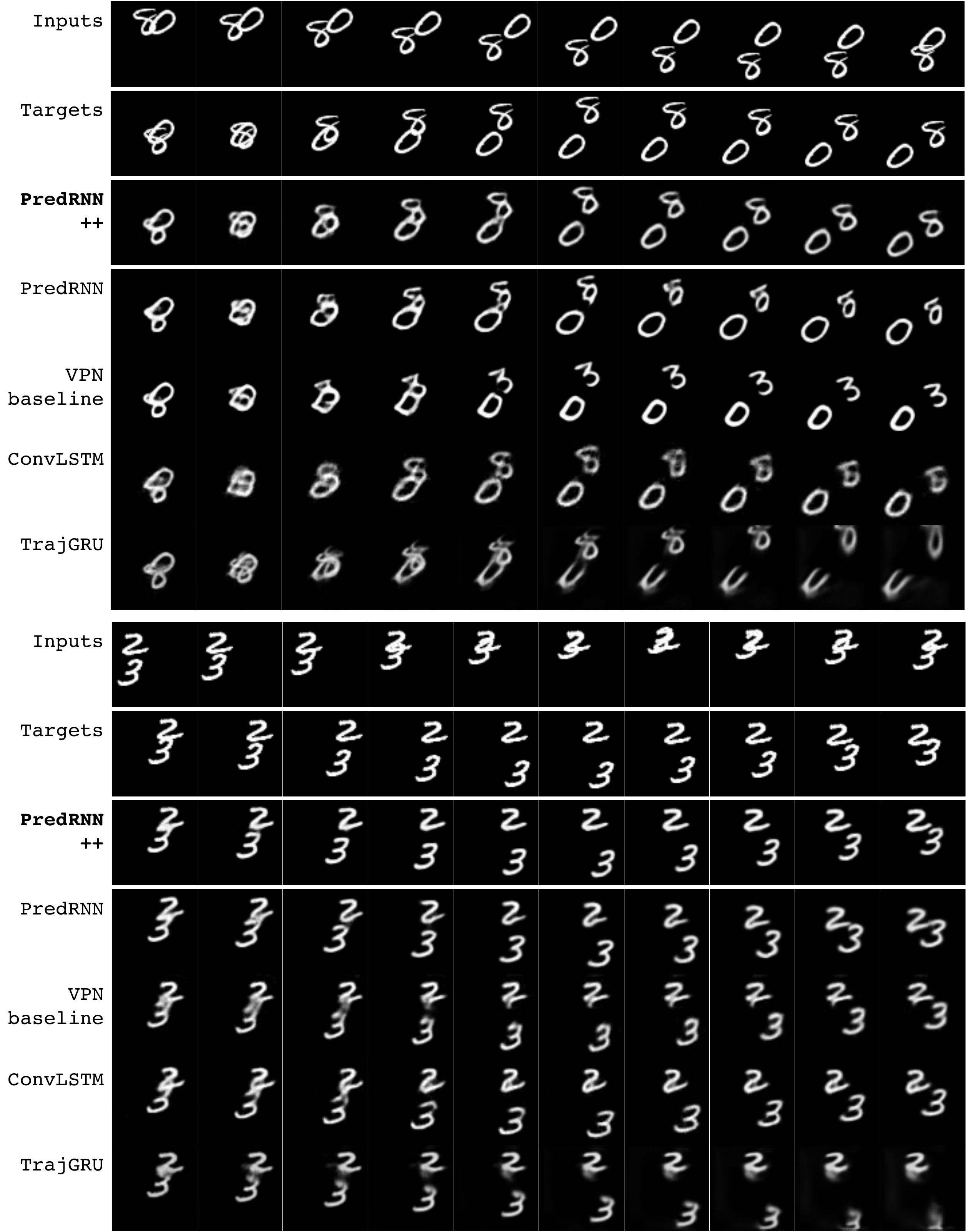}
\caption{Two prediction examples respectively with entangled digits in the input or output frames on Moving MNIST-2 test set.}
\label{fig:mnist_result}
\end{figure}

\begin{figure}[htb]
\centering
\subfigure[MNIST-2]{
\includegraphics[width=0.47\columnwidth]{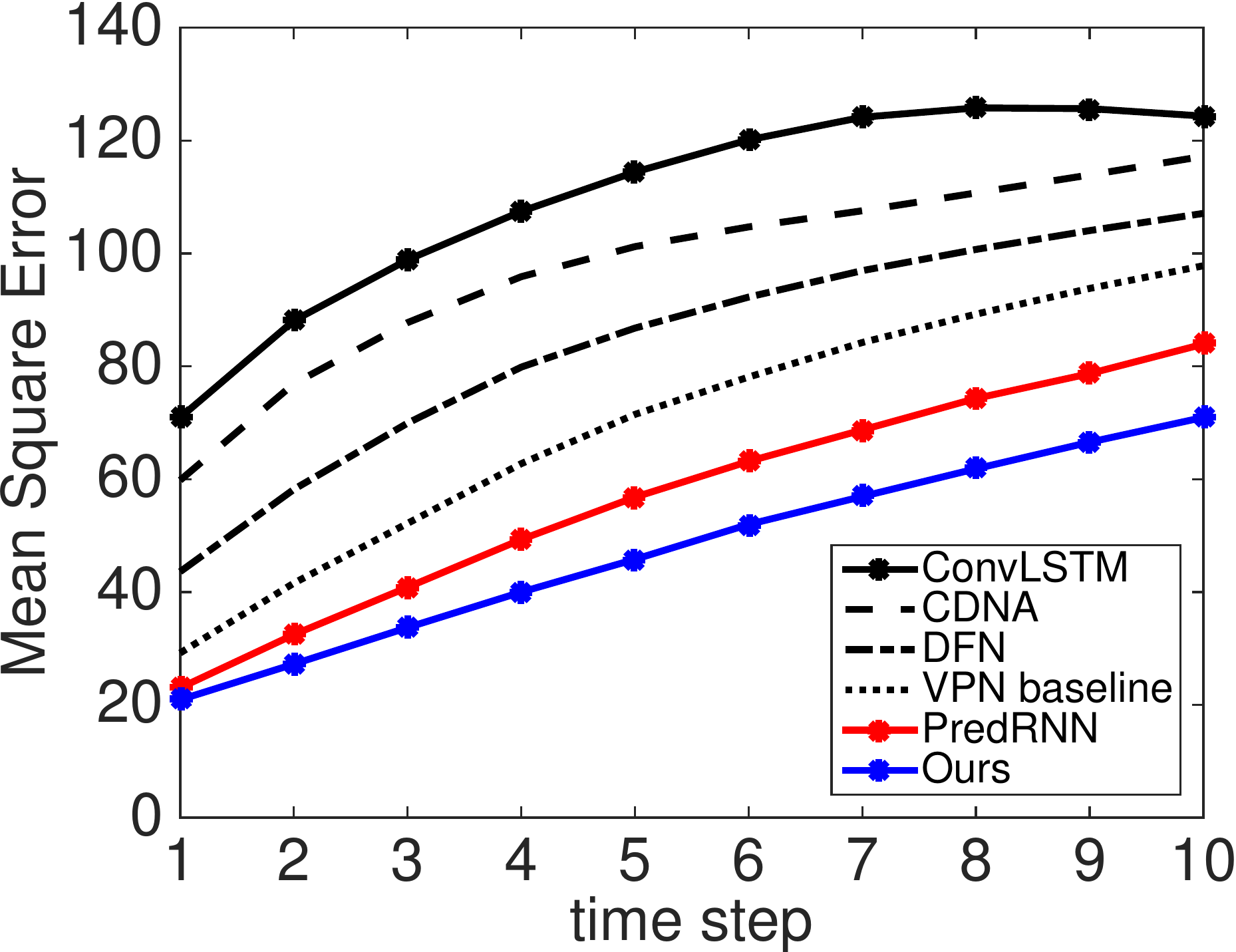}
\label{fig:mnist2_frame_mse}
}
\subfigure[MNIST-3]{
\includegraphics[width=0.47\columnwidth]{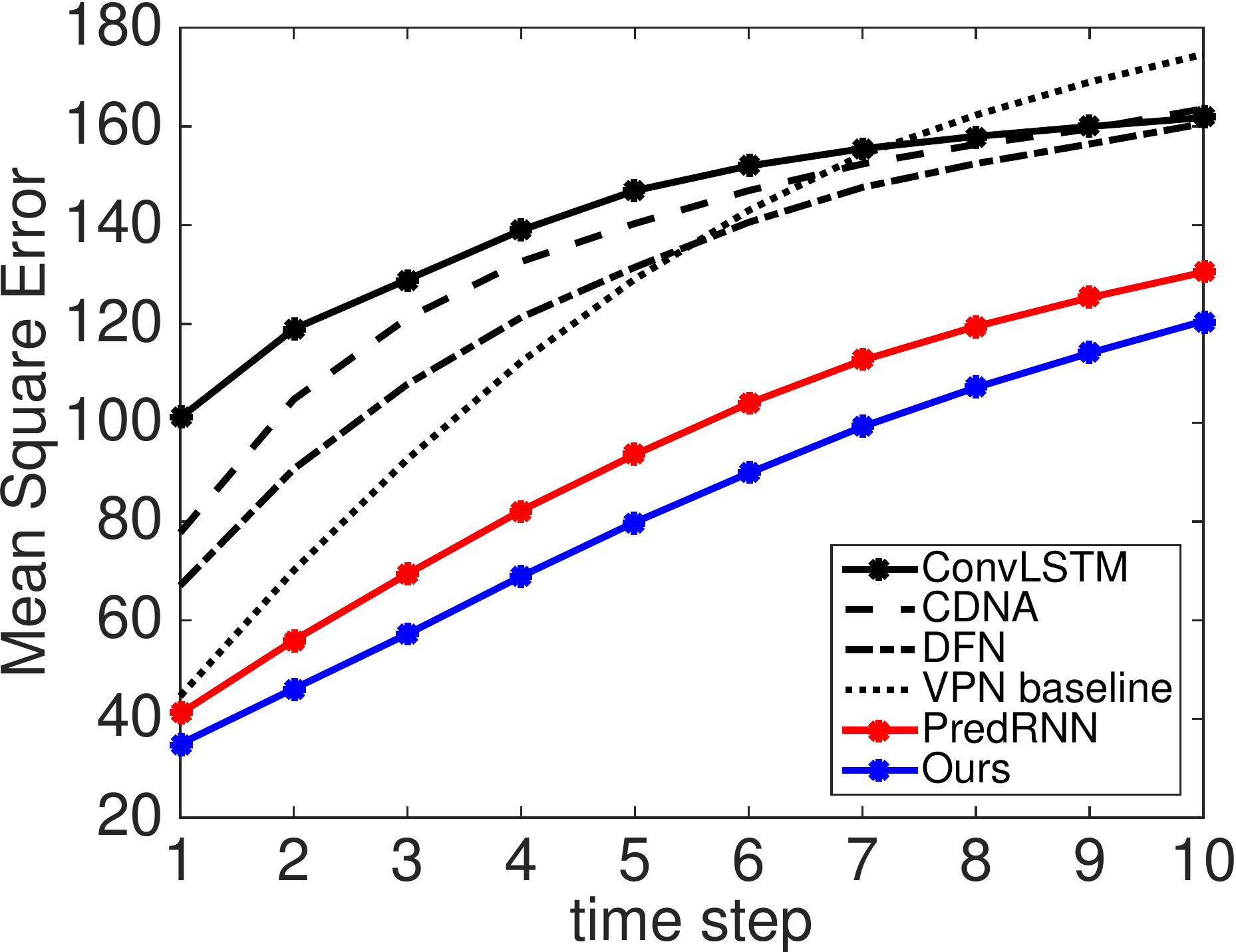}
\label{fig:mnist3_frame_mse}
}
\caption{Frame-wise MSE over the test sets. Lower curves denote higher prediction quality. All models are trained on MNIST-2.}
\label{fig:mnist_frame_mse}
\end{figure}

\paragraph{Results} 
To evaluate the performance of our model, we measure the per-frame structural similarity index measure (SSIM) \cite{Wang2004Image} and the mean square error (MSE). SSIM ranges between -1 and 1, and a larger score indicates a greater similarity between the generated image and the ground truth image. Table \ref{tab:mnist_mse} compares the state-of-the-art models using these metrics. In particular, we include the baseline version of the VPN model \cite{Kalchbrenner2016Video} that generates each frame in one pass. Our model outperforms the others for predicting the next 10 frames. In order to approach its temporal limit for high-quality predictions, we extend the predicting time horizon from 10 to 30 frames. Even though our model still performs the best in this scenario, it begins to generate increasingly more blurry images due to the inherent uncertainty of the future. Hereafter, we only discuss the 10-frame experimental settings.

Figure \ref{fig:mnist_frame_mse} illustrates the frame-wise MSE results, and lower curves denote higher prediction accuracy. For all models, the quality of the generated images degrades over time
. Our model yields a smaller degradation rate, indicating its capability to overcome the long-term information loss and learn skip-frame video relations with the gradient highway. 


In Figure \ref{fig:mnist_result}, we show examples of the predicted frames. With causal memories, our model makes the most accurate predictions of digit trajectories. We also observe that the most challenging task in future predictions is to maintain the shape of the digits after occlusion happens. This scenario requires our model to learn from previously distant contexts. For example, in the first case in Figure \ref{fig:mnist_result}, two digits entangle with each other at the beginning of the target future sequence. Most prior models fail to preserve the correct shape of digit ``8'', since their outcomes mostly depend on high level representations at nearby time steps, rather than the distant previous inputs (please see our afterwards gradient analysis). Similar situations happen in the second example, all compared models present various but incorrect shapes of digit ``2'' in predicted frames, while PredRNN++ maintains its appearance. It is the gradient highway architecture that enables our approach to learn more {disentangled representations} and predict both correct shapes and trajectories of moving objects.

\paragraph{Ablation Study} 
As shown in Table \ref{tab:mnist_mse}, it is beneficial to use causal LSTMs in place of ST-LSTMs, improving the SSIM score of PredRNN from $0.867$ to $0.882$. It proves the superiority of the cascaded structure over the simple concatenation in connecting the spatial and temporal memories. As a control experiment, we swap the positions of spatial and temporal memories in causal LSTMs. This structure (the spatial-to-temporal variant) outperforms the original ST-LSTMs, with SSIM increased from $0.867$ to $0.875$, but yields a lower accuracy than using standard causal LSTMs.

Table \ref{tab:mnist_mse} also indicates that the gradient highway unit (GHU) cooperates well with both ST-LSTMs and causal LSTMs. It could boost the performance of deep transition recurrent models consistently.
In Table \ref{tab:ablation}, we discuss multiple network variants that inject the GHU into different slots between causal LSTMs. It turns out that setting this unit right above the bottom causal LSTM performs best. In this way, the GHU could select the importance of the three information streams: the long-term features in the highway, the short-term features in the deep transition path, as well as the spatial features extracted from the current input frame.

\begin{table}[htb]
\caption{Ablation study: injecting the GHU into a 4-layer causal LSTM network. The slot of the GHU is positioned by the indexes ($k_1,k_2$) of the causal LSTMs that are connected with it.}
\label{tab:ablation}
\vskip 0.15in
\centering
\begin{small}
\begin{sc}
\begin{tabular}{lccc}
\toprule
Location & $k_1,k_2$ & SSIM & MSE \\
\midrule
Bottom (PredRNN++) 	& 1,2 &	\textbf{0.898} & \textbf{46.5} \\
Middle	  			& 2,3 & 0.894 & 48.1 \\
Top 				& 3,4 & 0.885 & 52.0 \\
\bottomrule
\end{tabular}
\end{sc}
\end{small}
\end{table}

\paragraph{Gradient Analysis} 

\begin{figure}[htb]
\centering
\subfigure[$\Vert\nabla_{\mathcal{X}_t}\mathcal{L}_{20}\Vert$]{
\includegraphics[height=0.368\columnwidth]{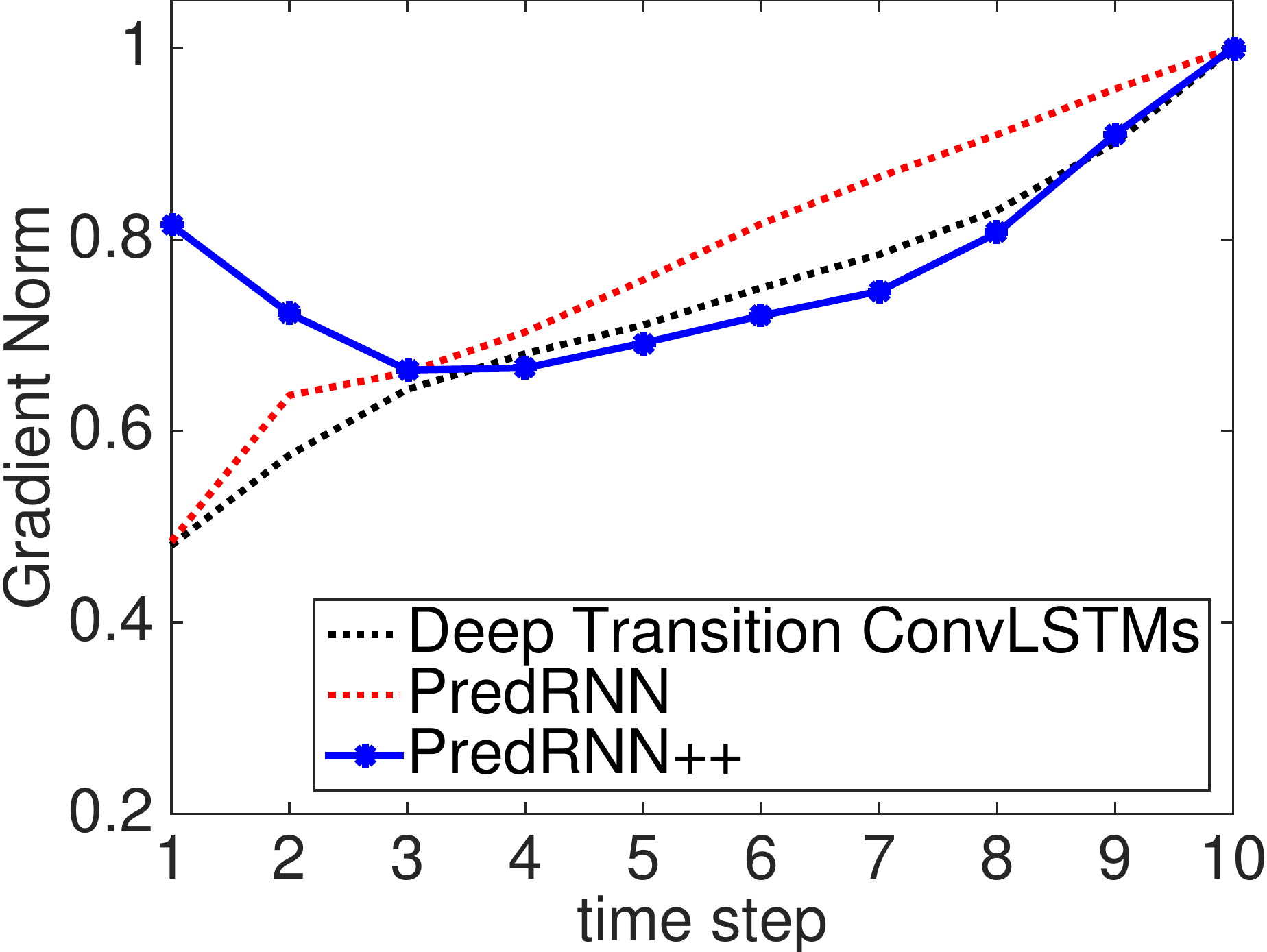}
\label{fig:inputs_gradient}
}
\subfigure[Occlusion Frequency]{
\includegraphics[height=0.368\columnwidth]{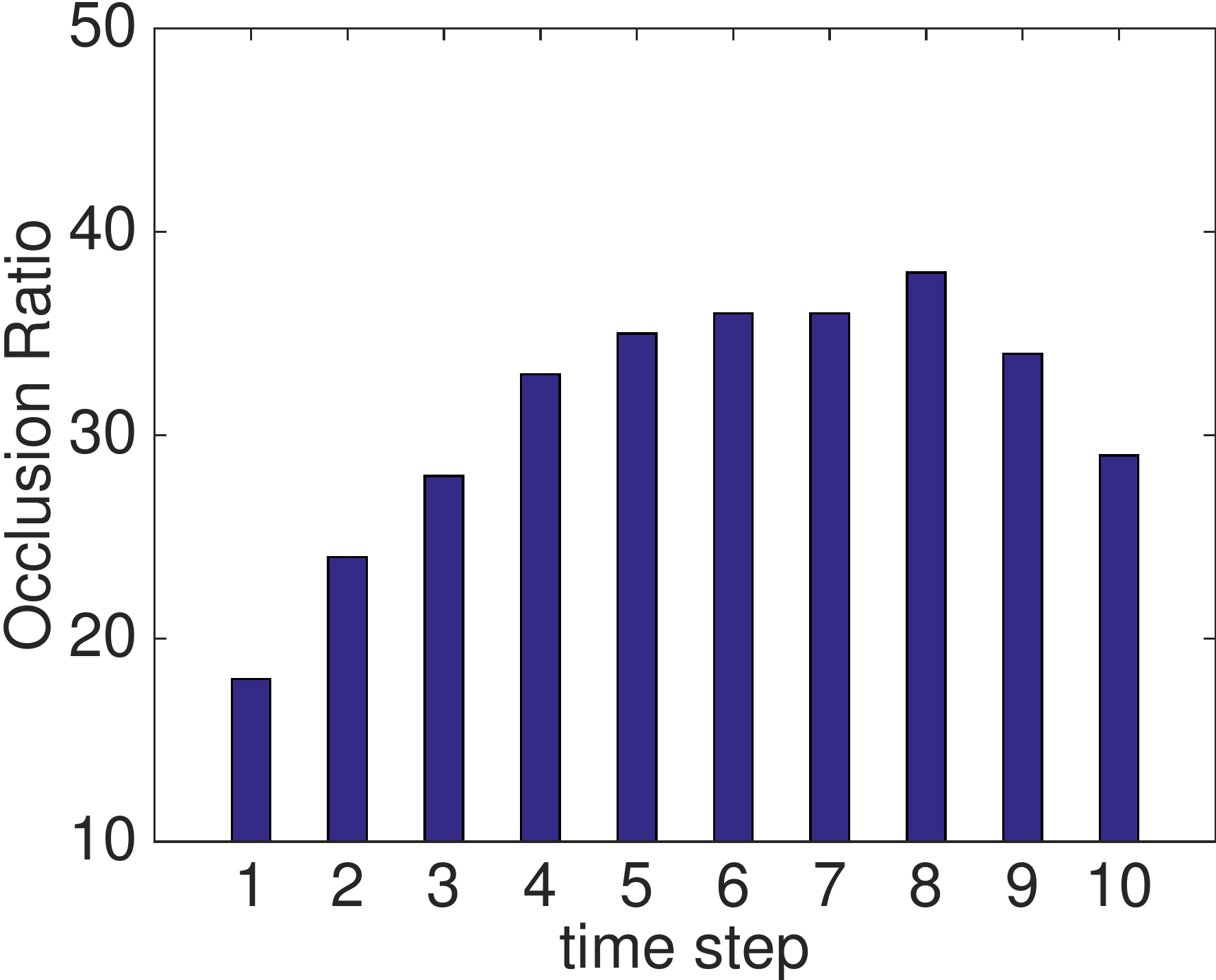}
\label{fig:overlap}
}
\caption{Gradient analysis: (a) The gradient norm of the loss function at the last time step with respect to each input frame, averaged over the test set. (b) The frequency of digits entangling in each input frame among $5,000$ sequences over the test set.}
\label{fig:gradient1}
\end{figure}

We observe that the moving digits are frequently entangled, in a manner similar to real-world occlusions. If digits get tangled up, it becomes difficult to separate them apart in future predictions while maintaining their original shapes. This is probably caused by the vanishing gradient problem that prevents the deep-in-time networks from capturing long-term frame relations. We evaluate the gradients of these models in Figure \ref{fig:inputs_gradient}. $\Vert\nabla_{\mathcal{X}_t}\mathcal{L}_{20}\Vert$ is the gradient norm of the last time-step loss function w.r.t. each input frame. Unlike other models that have gradient curves that steeply decay back in time, indicating a severe vanishing gradient problem, our model has a unique bowl-shape curve, which shows that it manages to ease vanishing gradients. We also observe that this bowl-shape curve is in accordance with the occlusion frequencies over time as shown in Figure \ref{fig:overlap}, which demonstrates that the proposed model manages to capture the long-term dependencies. 



\begin{figure*}[htb]
\centering
\subfigure[Deep Transition ConvLSTMs]{
\includegraphics[height=0.399\columnwidth]{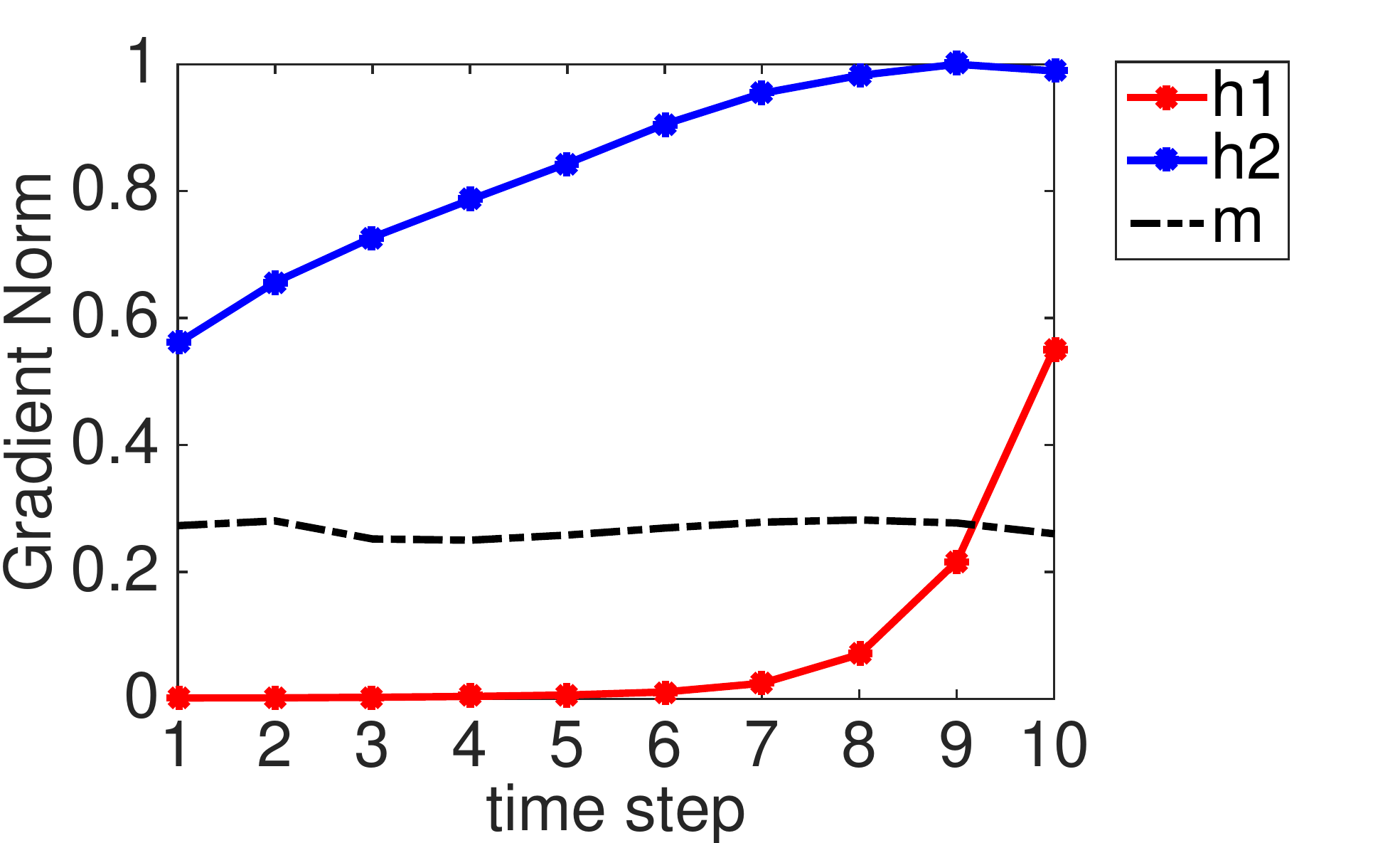}
\label{fig:convlstm_gradient}
}
\subfigure[PredRNN]{
\includegraphics[height=0.399\columnwidth]{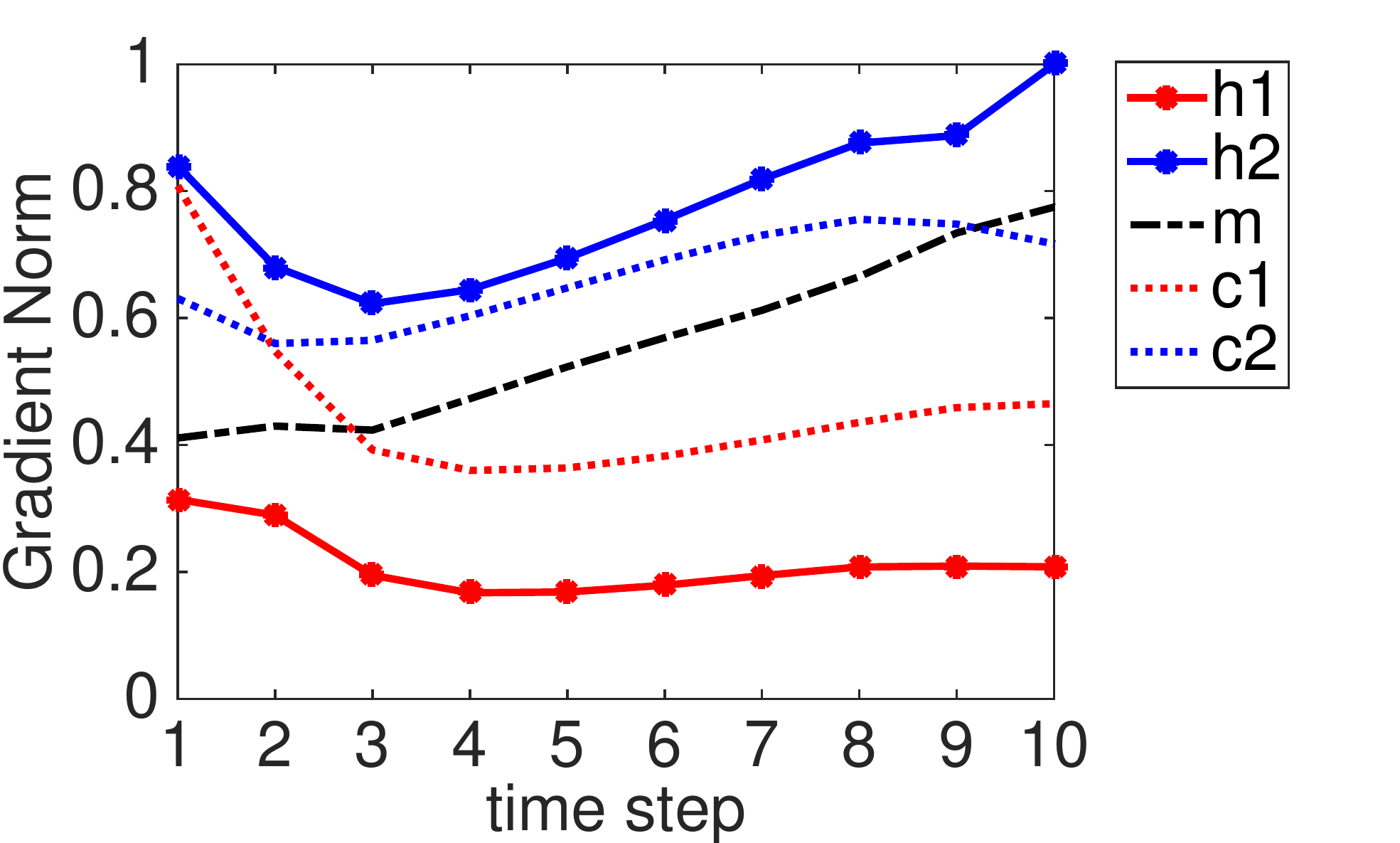}
\label{fig:predrnn_gradient}
}
\subfigure[PredRNN++]{
\includegraphics[height=0.399\columnwidth]{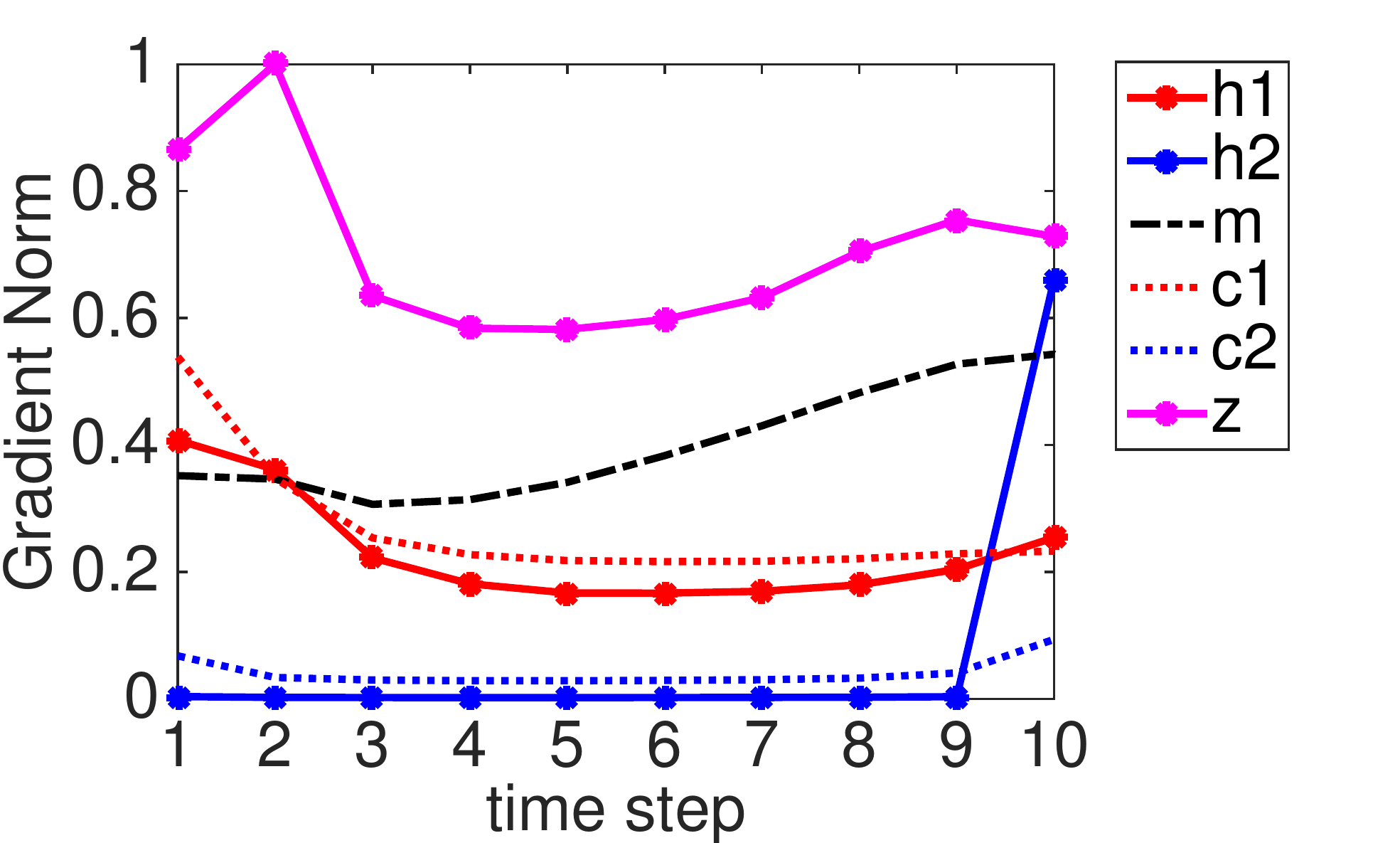}
\label{fig:predrnn_plus_gradient}
}
\caption{The gradient norm of the loss function at the last time step, $\mathcal{L}_{20}$, with respect to intermediate activities in the encoder, including hidden states, temporal memory states and the spatial memory states: $\Vert\nabla_{\mathcal{H}_t^k}\mathcal{L}_{20}\Vert$, $\Vert\nabla_{\mathcal{C}_t^k}\mathcal{L}_{20}\Vert$, $\Vert\nabla_{\mathcal{M}_t}\mathcal{L}_{20}\Vert$.}
\label{fig:gradient2}
\end{figure*}

Figure \ref{fig:gradient2} analyzes by what means our approach eases the vanishing gradient problem, illustrating the absolute values of the loss function derivatives at the last time step with respect to intermediate hidden states and memory states: $\Vert\nabla_{\mathcal{H}_t^k}\mathcal{L}_{20}\Vert$, $\Vert\nabla_{\mathcal{C}_t^k}\mathcal{L}_{20}\Vert$, and $\Vert\nabla_{\mathcal{M}_t}\mathcal{L}_{20}\Vert$. The vanishing gradient problem leads the gradients to decrease from the top layer down to the bottom layer. For simplicity, we analyze recurrent models consisting of $2$ layers. In Figure \ref{fig:convlstm_gradient}, the gradient of $\mathcal{H}_t^1$ vanishes rapidly back in time, indicating that previous true frames yield negligible influence on the last frame prediction. With temporal memory connections $\mathcal{C}_t^1$, the PredRNN model in Figure \ref{fig:predrnn_gradient} provides the gradient a shorter pathway from previous bottom states to the top. As the curve of $\mathcal{H}_t^1$ arises back in time, it emphasizes the representations of the more correlated hidden states. In Figure \ref{fig:predrnn_plus_gradient}, the gradient highway states $\mathcal{Z}_t$ hold the largest derivatives while $\Vert\nabla_{\mathcal{H}_t^2}\mathcal{L}_{20}\Vert$ decays steeply back in time, indicating that gradient highway stores long-term dependencies and allows causal LSTMs to concentrate on short-term frame relations. By this means, PredRNN++ disentangles video representations in different time scales with different network components, leading to more accurate predictions.

\subsection{KTH Action Dataset}

The KTH action dataset \cite{Sch2004Recognizing} contains $6$ types of human actions (walking, jogging, running, boxing, hand waving and hand clapping) in different scenarios: indoors and outdoors with scale variations or different clothes. Each video clip has a length of four seconds in average and was taken with a static camera in $25$ fps frame rate. 

\paragraph{Implementation} 
The experimental setup is adopted from \cite{Villegas2017Decomposing}: videos clips are divided into a training set of $108,717$ and a test set of $4,086$ sequences. Then we resize each frame into a resolution of $128\times128$ pixels. We train all of the compared models by giving them $10$ frames and making them generate the subsequent $10$ frames. The mini-batch size is set to $8$ and the training process is terminated after $200,000$ iterations. At test time, we extend the prediction horizon to $20$ future time steps.

\paragraph{Results} 

\begin{figure}[htb]
\vskip 0.15in
\centering
\includegraphics[width=\columnwidth]{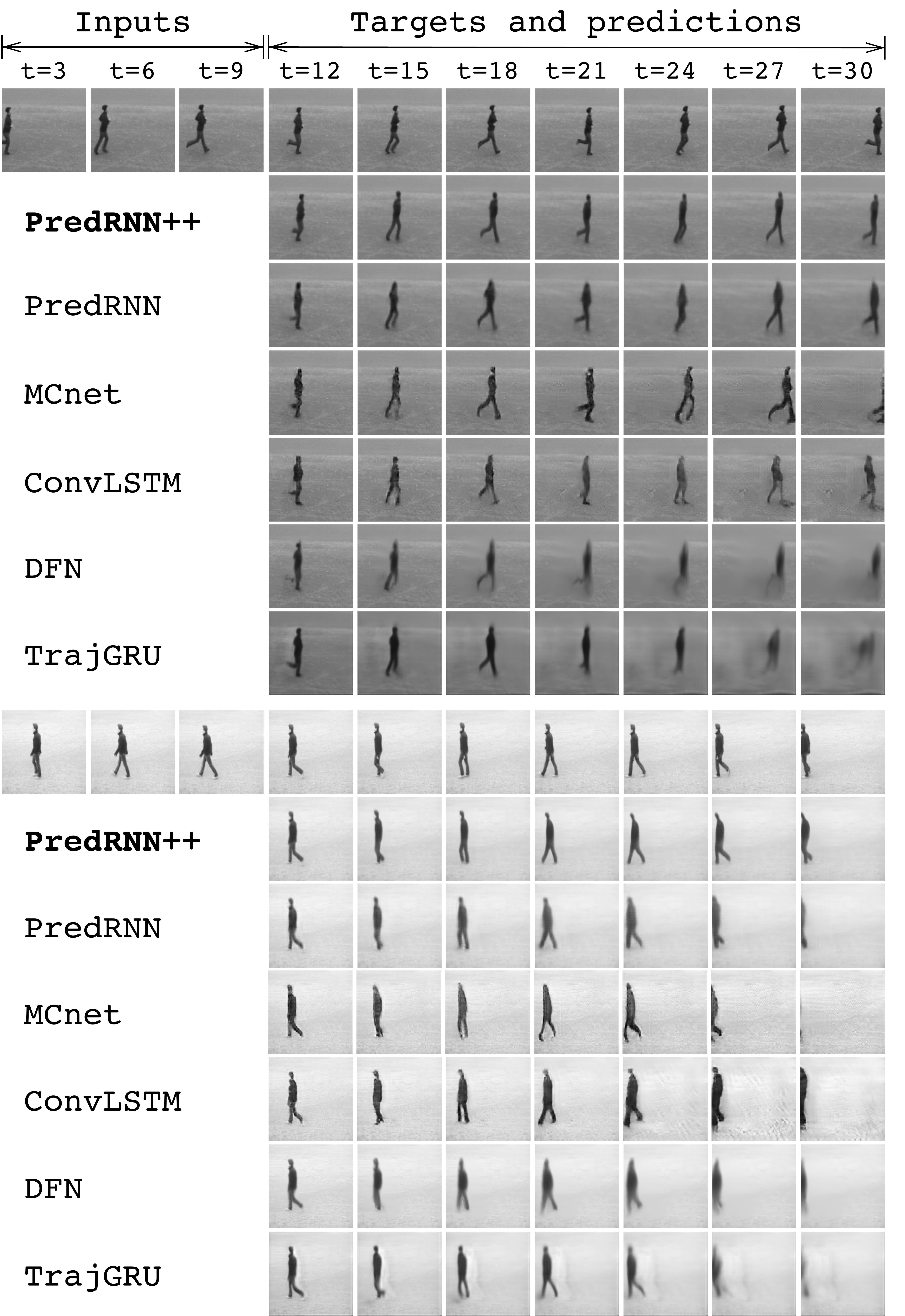}
\caption{KTH prediction examples. We predict 20 frames into the future by observing 10 frames. Frames are shown at a three frames interval. It is worth noting that these two sequences were also presented in \cite{Villegas2017Decomposing}.}
\label{fig:kth_results}
\end{figure}

Although few occlusions exist due to monotonous actions and plain backgrounds, predicting a longer video sequence accurately is still difficult for previous methods, probably resulting from the vanishing gradient problem. The key to this problem is to capture long-term frame relations. In this dataset, it means learning human movements that are performing repeatedly in the long term, such as the swinging arms and legs when the actor is walking (Figure \ref{fig:kth_results}). 

\begin{figure}[htb]
\vskip 0.15in
\centering
\subfigure[Frame-wise PSNR]{
\includegraphics[width=0.47\columnwidth]{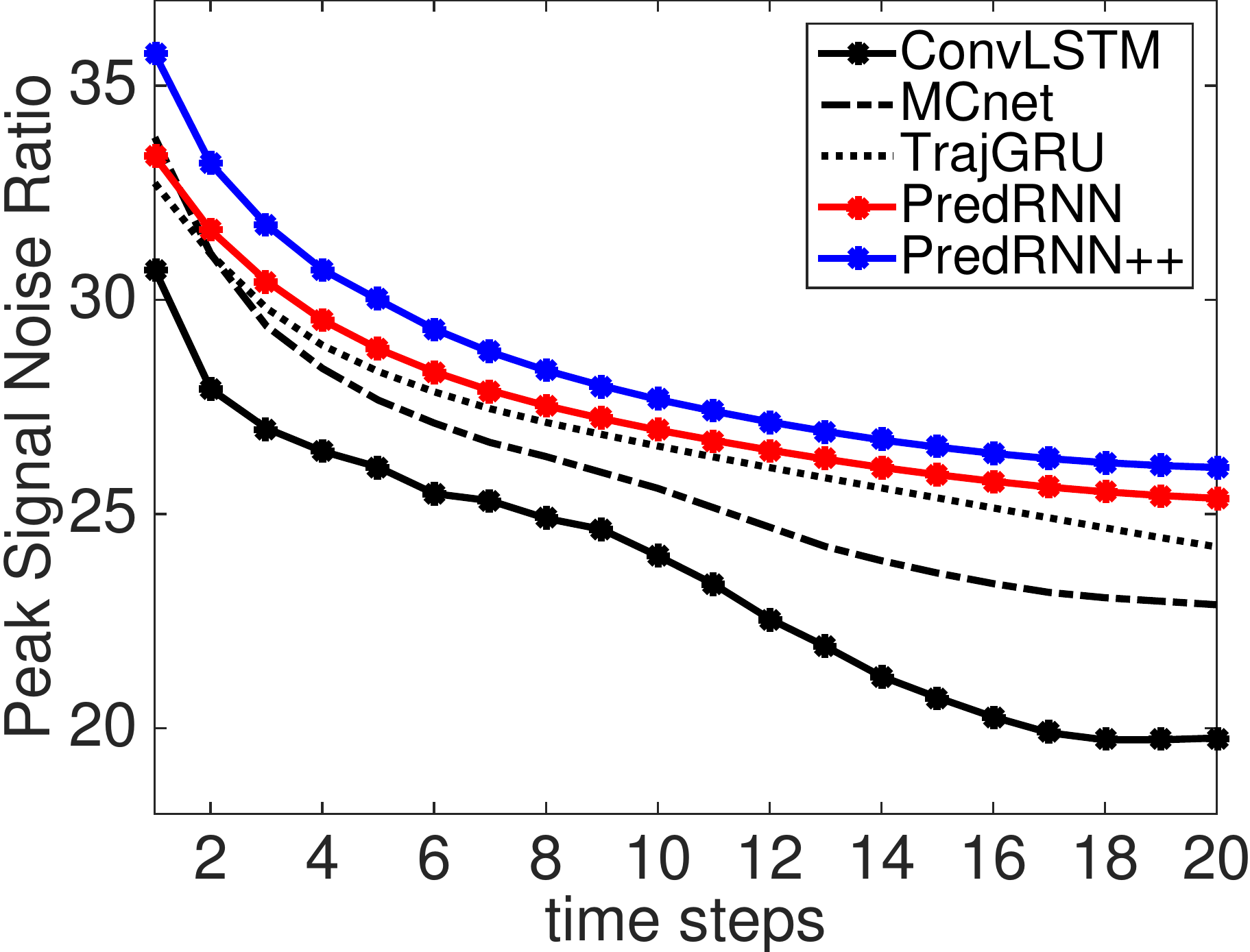}
}
\subfigure[Frame-wise SSIM]{
\includegraphics[width=0.47\columnwidth]{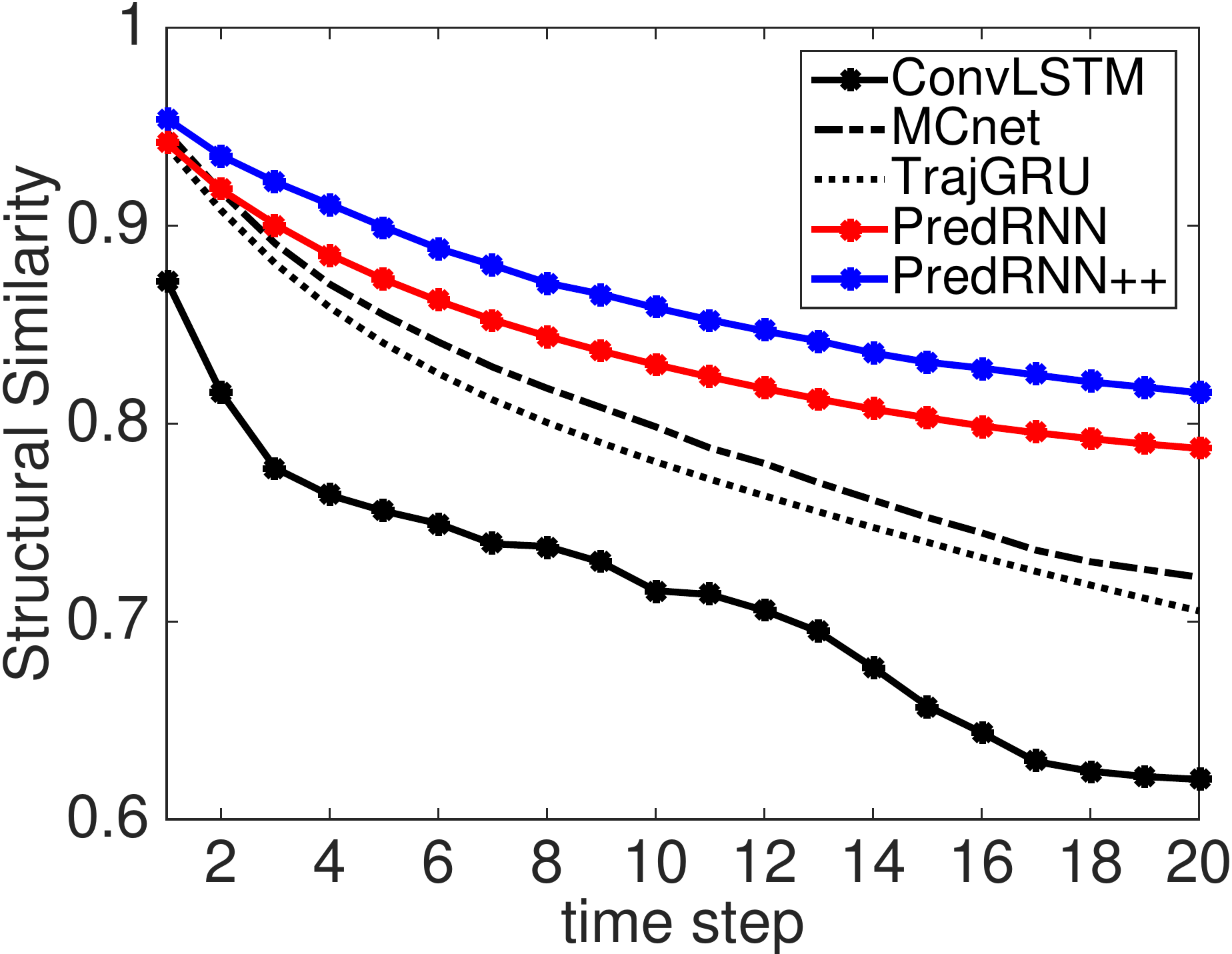}
}
\caption{Frame-wise PSNR and SSIM comparisons of different models on the KTH test set. Higher curves denote better results.}
\label{fig:kth_frame}
\end{figure}

We use quantitative metrics PSNR (Peak Signal to Noise Ratio) and SSIM to evaluate the predicted video frames. PSNR emphasizes the foreground appearance, and a higher score indicates a greater similarity between two images. Empirically, we find that these two metrics are complementary in some aspects: PSNR is more concerned about pixel-level correctness, while SSIM is also sensitive to the difference in image sharpness. In general, both of them need to be taken into account to assess a predictive model. Table \ref{tab:kth_comapre} evaluates the overall prediction quality. For each sequence, the metric values are averaged over the 20 generated frames. Figure \ref{fig:kth_frame} provides a more specific frame-wise comparison. Our approach performs consistently better than the state of the art at every future time step on both PSNR and SSIM. These results are in accordance with the quantitative examples in Figure \ref{fig:kth_results}, which indicates that our model makes relatively accurate predictions about the human moving trajectories and generates less blurry video frames.

\begin{table}[htb]
\caption{A quantitative evaluation of different methods on the KTH human action test set. These metrics are averaged over the 20 predicted frames. A higher score denotes a better prediction quality. }
\label{tab:kth_comapre}
\vskip 0.15in
\centering
\begin{small}
\begin{sc}
\begin{tabular}{lcc}
\toprule
Model & PSNR & SSIM \\
\midrule
ConvLSTM \cite{shi2015convolutional} & 23.58 & 0.712 \\
TrajGRU \cite{shi2017deep} 			 & 26.97 & 0.790 \\
DFN \cite{de2016dynamic} 			 & 27.26 & 0.794 \\
MCnet \cite{Villegas2017Decomposing} & 25.95 & 0.804 \\
PredRNN \cite{wang2017predrnn} 	 	 & 27.55 & 0.839 \\
\textbf{PredRNN++} 	& \textbf{28.47} & \textbf{0.865} \\
\bottomrule
\end{tabular}
\end{sc}
\end{small}
\end{table}

We also notice that, in Figure \ref{fig:kth_frame}, all metric curves degrade quickly for the first 10 time steps in the output sequence. But the metric curves of our model declines most slowly from the $10^\text{th}$ to the $20^\text{th}$ time step, indicating its great power for capturing long-term video dependencies. It is an important characteristic of our approach, since it significantly declines the uncertainty of future predictions. For a model that is deep-in-time but without gradient highway, it would fail to remember the repeated human actions, leading to an incorrect inference about future moving trajectories. In general, this ``amnesia'' effect would result in diverse future possibilities, eventually making the generated images blurry. Our model could make future predictions more deterministic.

\section{Conclusions}

In this paper, we presented a predictive recurrent network named PredRNN++, towards a resolution of the spatiotemporal predictive learning dilemma between deep-in-time structures and vanishing gradients. To strengthen its power for modeling short-term dynamics, we designed the causal LSTM with the cascaded dual memory structure. To alleviate the vanishing gradient problem, we proposed a gradient highway unit, which provided the gradients with quick routes from future predictions back to distant previous inputs. By evaluating PredRNN++ on a synthetic moving digits dataset with frequent object occlusions, and a real video dataset with periodic human actions, we demonstrated that it is able to learning long-term and short-term dependencies adaptively and obtain state-of-the-art prediction results.

\section{Acknowledgements}
This work is supported by National Key R\&D Program of China (2017YFC1502003), NSFC through grants 61772299, 61672313, 71690231, and NSF through grants IIS-1526499, IIS-1763325, CNS-1626432.

\nocite{langley00}

\bibliography{YWang}
\bibliographystyle{icml2018}

\end{document}